\documentclass[10pt, a4paper]{article}

\usepackage{multirow}
\usepackage[final]{lrec2026} 

\title{Text Simplification with Sentence Embeddings}

\name{Matthew Shardlow} 

\address{Manchester Metropolitan University \\
         m.shardlow@mmu.ac.uk\\}

\abstract{
Sentence embeddings can be decoded to give approximations of the original texts used to create them. We explore this effect in the context of text simplification, demonstrating that reconstructed text embeddings preserve complexity levels. We experiment with a small feed forward neural network to effectively learn a transformation between sentence embeddings representing high-complexity and low-complexity texts. We provide comparison to a Seq2Seq and LLM-based approach, showing encouraging results in our much smaller learning setting. Finally, we demonstrate the applicability of our transformation to an unseen simplification dataset (MedEASI), as well as datasets from languages outside the training data (ES,DE).  We conclude that learning transformations in  sentence embedding space is a promising direction for future research and has potential to unlock the ability to develop small, but powerful models for text simplification and other natural language generation tasks.
 \\ \newline  }

\begin{document}

\maketitleabstract

\section{Introduction}

We investigate the degree to which sentence embeddings can be used within the context of the text simplification task. Text simplification is a monolingual translation task in which complex sentences are converted into simpler alternatives \cite{saggion2017automatic}. In the data-driven paradigm the notions of complexity, simplicity and the mode of transition between them is governed by large parallel datasets containing examples of simplifications \cite{jiang-etal-2020-neural}. 

Sentence embeddings provide a distributional representation of the semantic contents of a text. For any given sequence of tokens $T = {t_1, t_2, ... t_n}$ we can apply some function $f(x)$ to transform $T$ into its corresponding embedding $E$ where $E$ is a fixed-size vector of real numbers. This gives the following formula describing the creation of some embedding E from any given sentence T:
\[
f(T) = E
\]
Recently, work has demonstrated that it is possible to decode the original content of a sentence based on the sentence embedding, giving the inverse formula $f^{-1}(y)$, or:
\[
f^{-1}(E) = T
\]
Such that $f^{-1}(y)$ reverses the effect of $f(x)$, leading to $f^{-1}(f(T)) = T'$, where $T \approx T'$

In this work, we investigate the potential of sentences reconstructed from sentence embeddings for the task of sentence simplification. Sentence simplification is a sub-domain of text simplification which seeks to transform some \textit{complex-original} sentence $T_c$ into a \textit{simple-target} sentence $T_s$. There are many approaches to sentence simplification, including data-driven approaches which rely on parallel corpora to learn sentence transformation operations.

In our study, we first investigate the degree to which reconstructed sentences preserve complexity levels by applying the $f^{-1}f(T_x) = T_x'$ transform and observing the differences between $T_x$ and the resulting $T_x'$. We then seek to learn some transformation $g(x)$ in sentence embedding space between the embeddings for complex $E_c$ and simple $E_s$ sentences:
\[
g(E_c) = E_s
\]
This then allows us to perform the following sequence of functions:
\[
f^{-1}(g(f(T_c))) = T_s
\]
whereby a sentence embedding $E_c$ is created for the complex sentence $T_c$ via the application of $f(x)$. Then, $E_c$ is transformed into $E_s$ via $g(x)$ and finally $T_s$ is given via the application of $f^{-1}(x)$ to $E_s$.

All code and data associated with this research is released via GitHub.\footnote{\url{https://github.com/mattshardlow/TextSimplificationWithSentenceEmbeddings}}

\section{Related Work}

Text simplification is an established field within natural language processing, seeking to create simpler representations of complex input texts \cite{ondov2022survey}. Typical approaches to text simplification range from lexical \cite{north2025deep}, and operations-based approaches \cite{dong-etal-2019-editnts} to sequence-to-sequence translation \cite{martin-etal-2022-muss} at the sentence and document level \cite{sun-etal-2021-document,blinova-etal-2023-simsum}. Text simplification corpora are available for Wikipedia-domain \cite{coster-kauchak-2011-simple,jiang-etal-2020-neural}, as well as many recent corpora in the medical domain \cite{trienes-etal-2022-patient,luo-etal-2022-benchmarking,horiguchi-etal-2024-evaluation,campillos-llanos-etal-2024-replace,joseph-etal-2023-multilingual}. Whereas text simplification resources are generally available in English, there is a growing trend towards the creation of text simplification resources for languages outside of English \cite{trienes-etal-2022-patient,horiguchi-etal-2024-evaluation,campillos-llanos-etal-2024-replace,joseph-etal-2023-multilingual,grabar-cardon-2018-clear,lee-vajjala-2022-neural,hartmann2020adaptaccao,dmitrieva-tiedemann-2021-creating}.

Whilst our task is primarily focussed on the simplification domain, the methodology we employ is that of sentence embeddings. Distributional semantic representations of texts at the lexical \cite{mikolov2013efficient,pennington-etal-2014-glove}, document \cite{10.1108/eb026526} and sentence \cite{reimers-gurevych-2019-sentence} levels are vital for modern deep learning based approaches to NLP. In particular, we make use of SONAR (SONAR stands for Sentence-level multimOdal and laNguage-Agnostic Representations) embeddings \cite{duquenne2023sonar} which provide a single common 1024-dimensional embedding space for sentences from diverse linguistic sources and modalities. SONAR embeddings are used in multiple downstream NLP tasks such as machine translation \cite{seamless2025joint} and concept modelling \cite{barrault2024large,dragunov2025sonar}.

\section{Methodology}

\subsection{Sentence Embeddings}
We make use of the SONAR embedding framework to create fixed-size representations of our source texts and also to decode embeddings into target texts. Sonar is multilingual (covering 200 languages) and multi-modal (trained on text and speech corpora), although we only use it in a mono-modal capacity in this work (text-embeddings only). SONAR embeddings and models are available via GitHub\footnote{\url{https://github.com/facebookresearch/SONAR/}} and we used these in the standard configuration following the example code.

SONAR relies on a Transformer encoder-decoder architecture \cite{vaswani2017attention}, which is initialised with pre-trained MT weights and optimised using parallel data according to a machine-translation objective, a denoising auto-encoding objective and a MSE loss objective. Sonar embeddings are created via mean-pooling of token-level encoder outputs in an auto-encoding setting. The encoder-decoder model was trained for 100K update steps across 200 languages against NLLB \cite{costa2022no}.

\subsection{Decoding Sentence Embeddings}

SONAR introduces random interpolation decoding, which takes the trained encoder-decoder model, freezes the encoder and fine-tunes the decoder weights on the random interpolation task. The task is defined as follows: 
\begin{quote}
  \textit{Given a bitext x, y, encode x and y with the frozen encoder, randomly draw z as a random interpolation of x and y embeddings, and learn to decode sentence embedding z into y.}
\end{quote}
This effectively combines the translation and auto-encoding tasks, allowing the decoder to produce new decoded texts based on a pooled representation of the encoder weights (the SONAR embedding).

\subsection{Datasets}

We additionally adopt a number of text simplification datasets from the text simplification literature for our task. These are..

\textbf{ASSET:} Comprising 2000 instances of validation data and 359 test examples, each with 10 reference simplifications. ASSET is widely used in text simplification evaluation in conjunction with the SARI metric. The simplifications in ASSET were collected via Crowdsourcing. Asset is in English only. \cite{alva-manchego-etal-2020-asset}

\textbf{WikiAuto:} An automatically aligned dataset comprising 488,332 parallel sentences from 138,095 article pairs extracted from English Wikipedia and Simple Wikipedia. WikiAuto is English only \cite{jiang-etal-2020-neural}.

\textbf{MedEASI:} A crowdsourced dataset of 1,979 parallel sentences from the medical domain. English only. We do not use this dataset for training, but we do evaluate against it for comparison to other systems \cite{basu2023med}.

\textbf{DEPlain (DE):}  A dataset in the German language consisting of news domain and web-domain texts which have been professionally simplified into easy-to-read German and manually aligned. We adopt 1883 instances of manually aligned web-text for testing only \cite{stodden-etal-2023-deplain}. 

\textbf{CLARA-MeD (ES):} A Spanish dataset in the medical domain with 1200 instances of parallel sentences extracted from 1040 announcements from the European Clinical Trials Register. All simplifications are produced manually at the sentence level\cite{campillos2022building}. 

\subsection{Evaluation Metrics}

\begin{description}
    \item[FKGL:] Flesch-Kincaid Reading Grade Level \cite{flesch1948new} is based on a readability formula that converts the numbers of words per sentence and syllables per word into a US educational grade level - indicating that a text is suitable for a reader at this level. 
    \item[ARI:] Similar to FKGL, the automated readability index \cite{senter1967automated} takes into account characters per word and words per sentence and is aligned to the US reading grade system.
    \item[CEFR-Prediction:] CEFR grade levels are an international standard in language learning, indicating the fluency of a learner. We make use of a sentence-level CEFR-prediction model from huggingface \texttt{AbdullahBarayan/ModernBERT-base-doc\_sent\_en-Cefr}, which is an instance of ModernBert \cite{modernbert}. To give a predicted CEFR level at the corpus level, we take the mean average of CEFR-level predictions for each corpus instance.
    \item[BLEURT:] An evaluation metric which is a regression model trained on human ratings of fluency and meaning preservation. BLEURT considers the reference sentences and the system outputs \cite{sellam-etal-2020-bleurt}.
    \item[SARI:] Widely used in text simplification evaluation and included here for completeness and comparison. SARI Score \cite{xu-etal-2016-optimizing} is the aggregated F1-score over the Addition, Keep and Delete operations as observed between the original source documents and references in comparison to the system outputs.
    \item[LENS:] A learnt metric for simplification quality \cite{maddela-etal-2023-lens}. LENS considers the source texts, system outputs and references and provides scores at the sentence level, which we average via the mean.
\end{description}

\section{Sentence Reconstruction} \label{sec:reconstruction}

\begin{table*}[ht]
    \centering
    \begin{tabular}{c|c|c|c|c|c|c|c}
                & Metric      & C      & C'     & $\Delta$C & S      & S'     & $\Delta$S\\
       \hline
       Asset    & FKGL        & 11.250 & 11.175 & -0.075      & 8.463  & 8.546  & 0.083 \\
                & ARI         & 11.458 & 11.375 & -0.083      & 8.149  & 8.232  & 0.083 \\
                & CEFR        & 2.447  & 2.451  &  0.004      & 2.096  & 2.137  & 0.041 \\
                & BLEURT      & ---    & ---    &  0.923      & ---    & ---    & 0.924 \\       
       \hline
       WikiAuto & FKGL        & 12.754 & 12.623 & -0.131      & 10.292 & 10.305 & 0.013 \\
                & ARI         & 12.896 & 13.035 &  0.138      & 9.849  & 10.095 & 0.246 \\
                & CEFR        & 3.092  & 3.119  &  0.027      & 2.717  & 2.794  & 0.077 \\
                & BLEURT      & ---    & ---    &  0.824      & ---    & ---    & 0.855 \\       
    \end{tabular}
    \caption{The results obtained when reconstructing Complex (C) or Simple (S) sentences for the ASSET Dataset and WikiAuto.}
    \label{tab:reconstruction_delta}
\end{table*}

In this section, we seek to answer the question of whether reconstructed sentences preserve the complexity level of the original sentences from which the corresponding embeddings were created. To investigate this question, we took the validation data from the Asset corpus (2000 instances) and a sample of the training data from the WikiAuto corpus (2000 instances). For Asset, each complex-original sentence has 10 simple-targets, but for this part of the study we only considered the first reference (index 0) for each complex-original sentence. Both datasets consist of complex-original sentences $C = {T_{c1}, T_{c2},...}$ and corresponding simple-target sentences $S = {T_{s1}, T_{s2},...}$, where each $T_{ci}$ has a corresponding $T_{si}$ which is the gold-standard simplified version of the complex-original sentence.

We then applied the transform $f^{-1}(f(x))$ to each instance of $C$ and $S$ to give $C' = = {T'_{c1}, T'_{c2},...}$ and $S'= {T'_{s1}, T'_{s2},...}$ respectively. We report results of this experiment in Table \ref{tab:reconstruction_delta}, where we show statistics for the complex-original sentences $C$, and their reconstructions $C'$ as well as the simple-target sentences $S$ and their reconstructions $S'$ in terms of FKGL, ARI and CEFR across both datasets. We additionally report BLEURT-score between C and C' and S and S' in each case.

The results in Table \ref{tab:reconstruction_delta} demonstrate that there is an improvement in reading ease between the complex-original sentences and the simple-target sentences. For example in the case of the ASSET dataset, the Flesch-Kincaid reading grade level (FKGL) for the complex-original sentences is 11.250, whereas the FKGL for the simple-target sentences is 8.463. This indicates that the simple sentences are around 2.787 reading grades apart. We should expect this as the simple-target sentences are designed to be simplified counterparts to the complex-original sentences. We also note the same effect for ARI. In WikiAuto, the reading grade level of the complex-original sentences is higher than that of those in Asset (12.754 vs. 11.250) indicating that the WikiAuto complex-original sentences are around 1.504 reading grades higher in complexity than those in ASSET. The same is true for the simple-target sentences, where those in WikiAuto are more than 2 reading grades higher than in Asset. There is however still a simplification effect in terms of FKGL between the complex-original sentences of WikiAuto (12.754) and the corresponding simple-target sentences (10.292), with a decrease of 2.462 reading grades.

We observe similar effects according to the average predicted CEFR levels for both datasets, corroborating the results derived from the two readability formulae employed. For Asset, the CEFR score is 2.447 for the complex-original data and 2.096 for the simple-targets, indicating that there is a simplification of around 0.351 of a CEFR level (i.e., mid-B1 to low-B1). For WikiAuto, there is a similar drop of 0.375 CEFR levels from 3.092 (low-B2) to 2.717 (mid-B1).

To determine the degree to which the reconstructed texts represent the original texts , we calculated the difference between the metrics for the reconstructed sentences ($T'$) and the original Sentences ($T$) as $\Delta T$. The reconstructed Texts (T') are very close to the original texts (T) in terms of complexity levels. For example in the Asset dataset, the FKGL of the complex-original texts is 11.250 and the FKGL of the reconstructed complex-original texts is 11.175 indicating a 0.075 drop in reading grade level. Similarly, there is a 0.083 increase in FKGL between the simple-target sentences and their reconstructions. The largest delta is in terms of ARI for the simple-target sentences of WikiAuto, where there is a 0.246 increase in ARI reading-level. This is around 10\% of the difference between the complex-original and simple-target sentences for WikiAuto.

To summarise our analysis, we demonstrate that there is a large difference between all metrics for the complex-original and simple-target sentences for both datasets. We also show that reconstructed sentences preserve the complexity level (FKGL, ARI, CEFR) and semantic content (BLEURT-score) of the original sentences. Finally, we also observe that the reconstructed complex-original and simple-target texts have a similar degree of difference to the original texts in terms of readability metrics, indicating that the degree of complexity change is also preserved. This allows us to confirm that reconstructed texts preserve complexity level.

\section{Sentence Simplification}

\begin{table}[t!]
    \centering
    \begin{tabular}{c|c|c|c}
            &           &        & Final \\
       K    & Params    & Epochs & Loss  \\\hline
       256  & 525,568   &  8200   & $4.712\times 10^{-5}$\\
       512  & 1,050,112 & 10000   & $4.302\times 10^{-5}$\\
       1024 & 2,099,200 & 10000   & $3.931\times 10^{-5}$ \\
       2048 & 4,197,376 &  9650   & $3.650\times 10^{-5}$\\
       4096 & 8,393,728 &  4800   & $3.618\times 10^{-5}$ \\
    \end{tabular}
    \caption{The results of training Neural Networks of various sizes to learn $g(x)$}
    \label{tab:NNLoss}
\end{table}

Given the findings in Section \ref{sec:reconstruction}, we hypothesise that it is possible to learn some function $g(x)$ that will represent the generic transformation between $E_c$ and $E_s$. To derive this function we make use of a multi-layer perceptron, trained through backpropagation on data from WikiAuto and ASSET with varying network sizes.  

For training our network, we rely on sentence-level data from Wiki-Auto and Asset. We make use 2000 instances of WikiAuto for validation data, the remainder of WikiAuto (n=486,332) for training and also the Asset validation data for training (n=2000). We test on the ASSET test set (n=357). For all datasets, the complex-original source data and the simple-target data was transformed into SONAR space via the SONAR encoder to give a 1024 dimensional representation of each original and target sentence.

\begin{table*}[ht]
    \centering
    \begin{tabular}{c|c|c|c|c|c|c}
             & Metric & C      & S      &  TSSE  & Seq2Seq & LLM    \\\hline
     Asset   & FKGL   & 11.668 &  8.154 &  8.303 & 7.126   & 9.022  \\
             & ARI    & 12.027 &  7.848 &  7.601 & 6.549   & 9.113  \\ 
             & CEFR   &  2.465 &  2.048 &  2.022 & 2.000   & 2.114  \\ \hline
     MedEASi & FKGL   & 13.736 & 11.476 & 10.953 & 11.530  & 10.532 \\
             & ARI    & 13.799 & 11.687 & 10.189 & 11.206  & 10.942\\
             & CEFR   &  3.490 &  3.100 &  3.103 & 3.173   & 2.907\\ 
    \end{tabular}
    \caption{Metrics demonstrating the measured complexity of the original texts (C), the simple references (S), our system (TSSE) and two baselines from the literature (Seq2Seq and LLM). We report for the ASSET Test set and MedEASi, both of which were unseen during model training.}
    \label{tab:assetReadability}
\end{table*}

\begin{table}[t]
    \centering
    \begin{tabular}{c|c|c|c|c}
               & Metric    & TSSE   & Seq2Seq & LLM  \\\hline
               & BLEURT    & 0.5909 & 0.7159  & 0.7314 \\
               & LENS      & 49.723 & 64.551  & 71.739 \\
       \parbox[t]{2mm}{\multirow{3}{*}{\rotatebox[origin=c]{90}{ASSET}}}       & SARI-Add  &  5.021 &  8.108  & 13.996 \\
               & SARI-Keep & 40.848 & 58.062  & 58.083 \\
               & SARI-Del  & 68.316 & 60.698  & 70.328 \\
               & SARI      & 38.061 & 42.290  & 47.469 \\\hline
       & BLEURT    & 0.4669 & 0.5722  & 0.5841 \\
               & LENS      & 27.473 & 37.809  & 47.342\\
     \parbox[t]{2mm}{\multirow{3}{*}{\rotatebox[origin=c]{90}{MedEASi}}}          & SARI-Add  &  2.111 &  2.578  &  5.834 \\
               & SARI-Keep & 26.460 & 44.058  & 39.940 \\
               & SARI-Del  & 76.568 & 58.812  & 74.374 \\
               & SARI      & 35.046 & 35.149  & 40.049 \\
    \end{tabular}
    \caption{Semantic metrics demonstrating the similarity of the sentences produced by our system (TSSE0 and 2 baselines (Seq2Seq and LLM) to the reference simplifications for ASSET test (10 refs) and MedEASi (1 ref).}
    \label{tab:assetSemantic}
\end{table}

\begin{table}[ht]
    \centering
    \begin{tabular}{c|c|c}
                    & Metric & TSSE \\\hline
      DEPlain (DE)  & BLEURT & 0.470 \\
                    & SARI   & 37.198 \\\hline
      ClaraMed (ES) & BLEURT & 0.266 \\
                    & SARI   & 25.996\\
    \end{tabular}
    \caption{The results of applying $g(x)$ for embeddings constructed from languages outside of the training data German (DE) and Spanish (ES). Only Bleurt and SARI are reported as these are language independent.}
    \label{tab:multilingual}
\end{table}

We configured a 2-layer fully-connected neural network using Pytorch, which consists of an input layer representing the embeddings (d=1024), a hidden layer of K nodes and an output layer of 1024 nodes. We experimented with K=[256,512,1024,2048,4096], allowing us to observe the effect of compressing (256,512) and expanding (2048,4096) the representation space. We used the Adam optimiser (learning rate = 0.001), with mean-squared error loss calculated between the model predictions and the reference embeddings. We allowed the network to run for 10K epochs, with checkpointing every 50 epochs and early stopping implemented for 5 successive checkpoints of loss increase on the validation data and rollback to the previous best model checkpoint. All experiments were conducted on an A100 GPU with 40GB of RAM via Google Colab.

We present results of training our neural network which approximates $g(x)$ in Table \ref{tab:NNLoss}, where we can observe that (a) the parameter count increases linearly with the size of K, (b) the parameter count is in the order of 0.5-8 million, which is much smaller than other models used for text simplification and (c) we are able to train a network in which the MSE loss calculated on the validation data decreases across epochs until the training ends (10K epochs) or the early stopping criteria is reached. The network with the lowest final loss has 4096 hidden nodes. We did not train any larger networks due to limitations on GPU RAM.

We then further test our neural network by running the full pipeline as evaluation $f^{-1}(g(f(C))) = S'$, where $S \approx S'$. We used the model with 2048 hidden parameters for this evaluation. In this setting, we take the complex-original test set of ASSET, encode it using the SONAR encoder, pass the resulting embeddings through the pre-trained neural network and then decode using the SONAR decoder. We term this approach `Text Simplification with Sentence Embeddings' or TSSE. The resulting decoded texts are then evaluated against the reference texts for ASSET using FKGL, ARI and CEFR-score in Table \ref{tab:assetReadability} and SARI score, BERTscore and LENS in Table \ref{tab:assetSemantic}. We include baseline systems which are taken from the \href{https://github.com/ZurichNLP/BLESS}{BLESS GitHub repository}. For both systems we made use of the generated system outputs that are available in the repository, however we did perform our own evaluation on each system output to ensure fairness and coverage over all metrics we report. The two baseline systems we use are a \textbf{Seq2Seq} baseline based on the \href{https://github.com/ZurichNLP/BLESS/blob/main/model_outputs_and_evals/openai-gpt-3.5-turbo/asset-test_asset-valid_p2_random_fs3_nr1_s723.jsonl}{MUSS} system \cite{martin-etal-2022-muss} and a \textbf{LLM} system based on \href{https://github.com/ZurichNLP/BLESS/blob/main/model_outputs_and_evals/openai-gpt-3.5-turbo/med-easi-test_med-easi-validation_p2_random_fs3_nr1_s723.jsonl}{GPT-3.5}. Full details of system configurations for these systems are available via BLESS \cite{kew-etal-2023-bless}.

\section{Multilingual Sentence Simplification}

Finally, we experiment within the SONAR embedding space to determine if a transform $g(x)$ learnt for English could be applicable to other languages. Effectively, we are testing the degree to which the representational space of the SONAR embeddings encodes cross-lingual simplification properties. To investigate this, we made use of the DEPlain and CLARA-MeD datasets in German and Spanish respectively. SONAR provides a multilingual encoder $f(x)$ and decoder $f^{-1}(x)$ which operates in the same space as for English. This meant that to implement the experiment in Spanish and German, we changed a single parameter referencing the language of the encoder/decoder.

For evaluation, we report only on SARI and BLEURT as these are language independent metrics. FKGL and ARI are both designed for English reading grade levels. The CEFR prediction and LENS models are both also trained on English sources only. SARI does not require a base model, instead relying on reference texts. BLEURT does make use of a base model, but this is multilingual, with coverage of German and Spanish.

For the experiment, we collected the complex-original texts for DEPlain and CLARA-MeD as well as the simple-target texts for each dataset. We encoded the original texts to give an embedding representation via SONAR, applied our model trained on English simplification data to the embeddings to give embeddings representing simplified versions of the original texts and finally decoded the resulting embeddings via SONAR using the appropriate language flag. We then compared the complex-original texts to the system outputs with respect to the simple-target texts via SARI and we also compared the system outputs to the simple-target texts via BLEURT score. 

The results in Table \ref{tab:multilingual} show mixed performance. Whereas the results for DEPlain show similar scores to those in English, the results for ClaraMed are much lower. The results for DEPlain are in line with those reported by the original study which report a SARI score of 34.828 on a fine-tuned MBart (see Table 5 in \cite{stodden-etal-2023-deplain}). It is notable that English and German are more ethnolinguistically similar than English and Spanish and it may be the case that the simplifications produced by $g(x)$ are more suitable for closely-related languages than those that are further apart, although we do not have sufficient data to fully investigate this claim. It is nonetheless encouraging to note that some form of simplification appears to have been produced for German, despite the model being trained exclusively on English data.

\begin{table*}[ht]
    \centering
    \begin{tabular}{c|p{6cm}|p{6cm}}
      Corpus   &  Original & System Output\\ \hline
      Asset    
      & In its pure form, dextromethorphan occurs as a white powder.
      & In pure form dextromethorphan is present as white powder.
      \\\cline{2-3}

      & At the Voyager 2 images Ophelia appears as an elongated object, the major axis pointing towards Uranus.
      & On the Voyager 2 images Ophelia appears as the large object with an axis that points to Uranus.
      \\\cline{2-3}

      & Today NRC is organised as an independent, private foundation.
      & Today NRC is organized as an independent, private foundation.
      \\\cline{2-3}

      & It is situated at the coast of the Baltic Sea, where it encloses the city of Stralsund.
      & It is located at the coast of Baltic Sea around the city of Stralsund.
      \\\cline{2-3}
      
     & He was also named 1982 "Sportsman of the Year" by Sports Illustrated.
     & He was named 1982 Sportsman of the Year by Sports Illustrated.

     \\ \hline \hline
     DEPlain

     & Der Generator kann rund sieben Watt erzeugen – um Mobiltelefone, Taschenlampen oder GPS-Geräte mit Energie zu speisen.
     & Der Generator kann bis zu 7 Watt Strom erzeugen. Er kann Blitzgeräte, GPS-Geräte und andere Geräte einsetzen.
     \\\cline{2-3}

     & Sie sollen die Natur dabei nah, intensiv und ursprünglich erleben.
     & Das Ziel ist es, die Natur zu erforschen, die Natur zu erforschen, die Natur zu erforschen.
     \\\cline{2-3}

     & Das Lebendige Museum Online (LeMO) ist ein Gang durch die deutsche Geschichte von der Mitte des 19. Jahrhunderts bis zur Gegenwart.
     & Das "Lebensmuseum" ist ein Museum aus dem 19. Jahrhundert.
     \\\cline{2-3}
     
     & So schnell kann man zu Fuß oder mit der Bahn keinen Rundgang machen.
     & Es ist nicht möglich, den Weg zu Fuß zu beschleunigen.
     \\\cline{2-3}

     & Google hat vor über einem Jahr 17 bedeutende Museen und Galerien im Internet unter "Art Project" für alle zugänglich gemacht.
     & Google hat vor 17 Jahren mehr als eintausend Kunstmuseen auf der ganzen Welt zugänglich gemacht.
     \\ \hline \hline

     MedEASi
     & Mujeres no postmenopáusicas (mínimo 12 meses) o quirúrgicamente estériles tienen que obtener un negativo en el test de embarazo de la visita de screening y al final del estudio.
     & Las mujeres que no son menopáusicas, o que tienen un embarazo prematuro, reciben una prueba de detección y una prueba de detección.
     \\\cline{2-3}
     
     & Mujeres en edad fértil que aún no están embarazadas en el momento de la entrada en el estudio y que no están dispuestas a abstenerse de tener relaciones sexuales con hombres o practicar métodos anticonceptivos apropiados hasta el día 140 del estudio
     & Las mujeres en edad fértil que no estén embarazadas están dispuestas a usar el anticonceptivo hasta los 40 días de embarazo.
     \\\cline{2-3}
     
     & Sujeto no podría recibir medicación que interfiera en la coagulación o la función plaquetaria en los 3 días previos a la primera dosis del fármaco del estudio o durante el periodo de tratamiento del estudio.
     & El paciente podría no recibir tratamiento con la droga durante el período de estudio.
     \\
     

    \end{tabular}
    \caption{Examples of system outputs for English, German and Spanish. The outputs demonstrate simplifications, as well as examples of hallucinations and decoding failures.}
    \label{tab:outputs}
\end{table*}

\section{Discussion}

We intended to investigate the degree to which sentence embeddings (specifically SONAR embeddings) can be used within the context of the text simplification task. 
Our results have demonstrated that complexity is preserved through encoding and decoding of sentence embeddings (See Table \ref{tab:reconstruction_delta} and that a small neural network (8M params, see Table \ref{tab:NNLoss}) is capable of learning a mapping between embeddings representing complex texts and embeddings representing simple texts. We further demonstrate that texts reconstructed from generated embeddings closely match the complexity level of reference simplifications across 2 datasets, including one out-of-domain dataset (see Table \ref{tab:assetReadability}). We also observed that our model underperforms compared to baseline models when using metrics that rely on simplified references --- indicating that there is still a gap in performance in the proposed approach. Finally, we report promising results on the potential to transfer the learnt mapping to other languages by making use of the shared multilingual embedding space, demonstrating encouraging metrics for German, but not for Spanish (see Table \ref{tab:multilingual}).

We have included several randomly sampled system outputs for the TSSE system for English, Spanish and German in Table \ref{tab:outputs}. These are presented with both the original complex sentence and the produced simplification. The simplification quality is typically higher for shorter sentences that do not contain too many named entities --- which is unsurprising given the degree of semantic compression that is taking place in the transformation to SONAR embeddings. We also note some failure cases of hallucinations in simplification, factual inaccuracies and at time neural text degeneration failures \cite{holtzman2019curious}. That being said many of the simplifications that were produced appear to preserve semantic content whilst transforming the texts at the syntactic and lexical levels to appear more simple.

We have not explicitly evaluated the degree to which simplified texts produced via sentence embeddings (or other means) are useful for a reader. We would note however that the metrics we have produced are aligned to human judgments and aligned to the simplified references, indicating that the texts that are produced are `simple' according to the definition of simplicity derived from the training corpora.

Our system is not intended as a new state-of-the-art for the text simplification task, but rather as an exploration of the possibilities of transformations within sentence embedding space as applied to text simplification. We are able to demonstrate the production of competitive simplifications with a very small model using a simple neural architecture. We expect that future work in this new exciting paradigm of natural language processing will allow for future improved results over those that we have produced here. 

\section*{Limitations}
A principal limitation on our work was the capacity to use high-spec GPU machines at training time. This limited our study to a maximum size of 4096 hidden nodes, and we suspect that a larger hidden layer or a more complex neural architecture may have been suitable for further improving the performance of our model. Additionally, this placed a limitation on how much data we were able to train our model against.

Our model is arguably also limited by the availability of high quality text simplification corpora. The largest such corpus for English is WikiAuto with 488,332 parallel sentences, however these are automatically aligned and do not conform to any definition of `simplified language' beyond that of guidelines provided to Simple Wikipedia editors. The provision of large-scale high-quality simplified corpora would greatly improve data-driven simplification approaches such as ours. 

Finally, our approach is fundamentally self-limited by the reliance on a 1024 embedding space. This creates a fixed-size representational space over which we have no control --- we did not retrain the encoder or decoder, instead only operating on the embeddings themselves. Retraining the encoder-decoder model for the sentence embeddings (e.g., to add additional languages), would be expensive and beyond the capacity of most researchers.

\section{Bibliographical References}\label{sec:reference}

\bibliographystyle{lrec2026-natbib}
\bibliography{lrec2026-example,anthology-1,anthology-2}


\end{document}